\documentclass{article}

\usepackage{microtype}
\usepackage{graphicx}
\usepackage{subcaption}
\usepackage{booktabs}
\usepackage{multirow}
\usepackage{xcolor}
\usepackage{pifont}

\usepackage{hyperref}

\usepackage[accepted]{icml2026}

\makeatletter
\renewcommand{\Notice@String}{Accepted at the ICML 2026 Workshop on
  Foundation Models for Structured Data (FMSD), Seoul, South Korea.
  Non-archival.}
\makeatother

\hypersetup{
  pdftitle={Towards Continuous-time Causal Foundation Models},
  pdfauthor={Dennis Thumm, Ruben Wiedemann, Ying Chen},
  pdfsubject={ICML 2026 Workshop on Foundation Models for Structured Data (FMSD)},
  pdfkeywords={Causal inference, Prior-Data Fitted networks, Time series,
    Stochastic differential equations, Foundation models}
}

\usepackage{amsmath}
\usepackage{amssymb}
\usepackage{mathtools}
\usepackage{amsthm}

\usepackage[capitalize,noabbrev]{cleveref}

\theoremstyle{plain}
\newtheorem{theorem}{Theorem}[section]

\theoremstyle{definition}
\newtheorem{definition}[theorem]{Definition}

\usepackage[disable]{todonotes}

\newcommand{\doop}{\mathrm{do}}
\newcommand{\Pa}{\mathrm{Pa}}

\newcommand{\Gcal}{\mathcal{G}}

\icmltitlerunning{Towards Continuous-time Causal Foundation Models}

\begin{document}

\twocolumn[
  \icmltitle{Towards Continuous-time Causal Foundation Models}

  \begin{icmlauthorlist}
    \icmlauthor{Dennis Thumm}{nus}
    \icmlauthor{Ruben Wiedemann}{icl}
    \icmlauthor{Ying Chen}{nusym}
  \end{icmlauthorlist}
  \icmlaffiliation{nus}{National University of Singapore}
  \icmlaffiliation{icl}{Imperial College London}
  \icmlaffiliation{nusym}{Department of Mathematics, Centre for Quantitative Finance, Risk Management Institute, National University of Singapore}
  \icmlcorrespondingauthor{Dennis Thumm}{dennis.thumm@u.nus.edu}

  \icmlkeywords{Causal inference, Prior-Data Fitted networks, Time series, Stochastic
    differential equations, Foundation models}

  \vskip 0.3in
]

\printAffiliationsAndNotice{}

\begin{abstract}
  Extending discrete-time causal Prior-data Fitted Networks for time series to continuous time invites writing the mechanism as a stochastic differential equation (SDE)---but if the SDE is integrated \emph{once per observation gap}, the trajectory law depends on when it is observed, and the prior remains a discrete-time Markov model in SDE clothing.
  We propose a precise continuity criterion---trajectory-law invariance to the observation schedule---together with a three-tier taxonomy (discrete; naive observation-grid integration; fine-grid integration with decoupled observation) and a construction realising the top tier on a random DAG with OU or small-MLP nonlinear drifts, irregular observation schedules, and hard / soft / time-varying interventions.
  A $2 \times 2$ encoder $\times$ integrator ablation, run independently on a linear and a nonlinear prior, finds fine-grid integration beats naive on 8/8 cells (sign-consistency $p < 1/256$) with the gap growing as the eval grid refines; the encoder axis is null with fine integration but time-aware-leading with naive.
  We release\footnote{\url{https://github.com/thummd/continuous-time-causal-pfn}} the prior and a preliminary zero-shot protocol on pharmacokinetic and physical-system data.
\end{abstract}

\section{Introduction}
\label{sec:intro}

Prior-Data Fitted networks (PFNs) \citep{muller2022transformers, hollmann2023tabpfn, nagler2023statistical} pre-train a transformer on datasets sampled from an analytic data-generating prior and then perform in-context inference at test time. 
In causal settings, Do-PFN \citep{robertson2025dopfn} and CausalFM \citep{ma2026foundation} have pushed this recipe to \emph{interventional} tabular prediction by training on synthetic structural causal models (SCMs) \citep{pearl2009causality}. Recent work extends causal PFNs to multivariate time series by sampling temporal SCMs (TSCMs) with lagged directed acyclic graphs (DAGs), nonlinear autoregressive mechanisms, and multiple intervention types \citep{thumm2026interventional}.

Existing temporal causal priors \citep{thumm2026interventional} are \emph{discrete-time}: the generating process steps on a regular integer grid and the lag structure is a stack of adjacency matrices indexed by integer offsets.
A natural response is to rewrite the mechanism as a stochastic differential equation (SDE) and let it run between observations.
But the devil is in the integration:
if the SDE is stepped \emph{once per observation gap} (Euler--Maruyama (EM) on the observation grid), the joint law of a trajectory depends on when it is observed, and the prior is still effectively a discrete-time Markov model in SDE clothing. The target domains that motivate continuous time---pharmacokinetic concentrations sampled at clinically chosen times \citep{boeckmann1994nonmem}, physical systems like the Causal Chamber \citep{gael2024causalchamber} with variable-delay events, and electronic health records with missing-at-random and missing-not-at-random gaps \citep{che2018grud,rubanova2019latentode}---are \emph{schedule-heterogeneous} and require more.

This paper takes a step back and asks what exactly a causal PFN prior must satisfy to be called continuous-time. Our contributions are:
\begin{enumerate}
  \setlength{\itemsep}{2pt}\setlength{\parskip}{0pt}
  \item A \textbf{precise criterion for continuous-time causal priors}
    (\Cref{sec:definition}): the joint law of a sampled trajectory must be invariant to the observation schedule. We give a three-tier taxonomy---discrete ($\Delta t \equiv 1$), naive observation-grid integration, and fine-grid integration with decoupled observation---that operationalises the criterion.
  \item A \textbf{construction that realises the top tier}
    (\Cref{sec:prior}): Ornstein--Uhlenbeck (OU) or small-Multilayer Perceptron (MLP) nonlinear drifts on a random DAG with optional hidden confounders and Markov  regime switches, irregular observation schedules, and hard / soft / time-varying interventions, all integrated on a fine grid and subsampled to the observation schedule.
  \item An \textbf{empirical $2\times2$ encoder $\times$ integrator ablation}
    (\Cref{sec:experiments}), run independently on a linear-OU prior and a nonlinear neural-drift prior (4 cells $\times$ 2 priors $\times$ single seed $\times$ 10\,k steps). The (B)-vs-(C) gap is positive on every encoder cell across three eval discretisations on each of the two priors (Tables \ref{tab:reg}--\ref{tab:substeps}, 8/8; sign-consistency $p < 1/256$ under no-effect null). The lead is smallest on the eval that matches the naive variant's training schedule and substep tier and grows when the eval moves to finer substeps. The encoder axis is null with fine integration; with naive integration the time-aware encoder leads on both priors---consistent with fine integration making the data-generating process approximately schedule-invariant and removing the need for explicit time-gap features.
\end{enumerate}
Real-data transfer (Theophylline, Warfarin, Causal Chamber) is preliminary and deferred to Appendix \ref{app:realdata};
the main body argues the continuity case on synthetic data where it can be measured cleanly.

\section{Background and Related Work}
\label{sec:related}

\paragraph{Causal PFNs.} Do-PFN \citep{robertson2025dopfn} and CausalFM
\citep{ma2026foundation} pre-train transformers on SCMs and estimate conditional interventional distributions in context on independent and identically distributed (i.i.d.) tabular data.
They do not address temporal dependencies.

\paragraph{Temporal interventional priors.} Only a handful of generators produce paired (observational, interventional) time-series data:
CAnDOIT \citep{castri2024candoit} restricts to hard interventions at known targets; TECDI/RealTCD \citep{li2023tecdi,li2024realtcd} handle soft or hard interventions in linear structural vector auto-regressive (SVAR) models; CaTSG \citep{xia2025catsg} approximates $\doop$-calculus with a learned diffusion model. The most recent CausalTimePrior framework \citep{thumm2026interventional} samples nonlinear autoregressive TSCMs with hard, soft, and time-varying interventions---but, like all of the above, on a discrete-time grid. We build directly on its lagged-DAG formulation \citep{boeken2024dscm} and replace the mechanism and schedule with continuous-time analogues.

\paragraph{Continuous-time dynamical ML and SDE causality.} Neural ODEs \citep{chen2018neural}, Neural SDEs \citep{kidger2021neural}, and latent-ODE models for irregular series \citep{rubanova2019latentode} demonstrate that continuous-time parameterisations can match or beat discrete ones on irregular data. Irregular-time attention \citep{shukla2021mtand, tashiro2021csdi} and time-series foundation models \citep{dooley2023forecastpfn, taga2025timepfn,moroshan2025tempopfn, xie2025cauker} ingest continuous timestamps but, to our knowledge, none target \emph{interventional} in-context prediction. Closest in spirit to our SDE-based prior, \citet{lorch2024stationary} \emph{learn} a single SDE whose stationary distribution captures interventional behaviour, dropping acyclicity.
Our goal is instead to \emph{sample} an analytically specified prior over SDE-driven TSCMs so a transformer can amortise causal inference across the family; the two approaches are complementary.

\section{Method}
\label{sec:method}

\subsection{What makes a causal prior continuous-time?}
\label{sec:definition}

Let $\mathcal{P}$ be a prior over (TSCM, trajectory) pairs, and let $\mathcal{P}_\tau$ denote the distribution of observations at schedule $\tau = (t_1 < \ldots < t_T)$.

\begin{definition}[Continuous-time causal prior]
\label{def:ct-prior}
$\mathcal{P}$ is \emph{continuous-time} if there exists a continuous-path stochastic process $X$ whose law is independent of $\tau$ and $\mathcal{P}_\tau$ is the law of $X|_\tau$. I.e.\ the observation schedule is pure measurement, not part of the TSCM.
\end{definition}

The definition partitions priors into three tiers:
(A)~\emph{discrete} ($\Delta t \equiv 1$), a VAR-style SCM \citep{thumm2026interventional} defined only on the integer grid and failing Definition \ref{def:ct-prior}
by construction;
(B)~\emph{naive continuous} (observation-grid integration), an SDE stepped once per observation gap $\Delta_i$---the joint kernel parameterises to a different Markov model as $\Delta_i$ varies, so the law depends on $\tau$;
(C)~\emph{continuous} (fine-grid integration), the SDE integrated on $\Delta_{\mathrm{fine}} \ll \min_i \Delta_i^{\mathrm{obs}}$ and subsampled to $\tau$, converging to the true SDE law as $\Delta_{\mathrm{fine}} \to 0$ \citep{kloeden1992numerical}. At any finite $\Delta_{\mathrm{fine}}$ tier (C) realises Definition \ref{def:ct-prior} only approximately, with $\|\mathcal{P}_\tau^{(C)} - \mathcal{P}_\tau^{(\mathrm{SDE})}\| \to 0$ as $\Delta_{\mathrm{fine}} \to 0$; we treat tier (C) as the practical realisation of the criterion.

Whether tiers (B) and (C) differ in practice depends on a stability
condition. The standard Euler--Maruyama update on a 1-D OU process
$dX = -\theta X\,dt + \sigma\,dW$ has amplification factor
$|1-\theta\Delta|$ per step and is mean-square stable only when
$\theta\Delta < 2$; on a prior that crosses this boundary, naive EM
produces exploding trajectories that pin training-target distributions
at their normalisation ceiling---a numerical-stability artefact rather
than a discretisation-bias signature.
Stability is necessary but not sufficient for naive $\approx$ fine: the leading per-step Euler--Maruyama bias against the exact Gaussian OU transition kernel is $O(\theta\Delta)$ on the variance, accumulating over the trajectory at the prior's typical $\theta\Delta \approx 0.3$. Eval-loss is partially robust to this transition-kernel bias---it scores predictive likelihood, not path-measure distance---so we expect a smaller but still detectable empirical gap, which \Cref{sec:experiments} confirms on both OU and neural priors. The construction (\Cref{sec:prior}) therefore pairs tier-(C) integration with a stability-respecting prior, and the ablation tests both axes.

\subsection{Construction of the continuous-time prior}
\label{sec:prior}

\paragraph{Graph sampling.} A sample from the prior draws $N \sim
\mathrm{Uniform}(3, N_{\max})$ variables and a DAG $\Gcal$ over them \citep{thumm2026interventional}. We
provide two graph samplers: (i) a named-structure sampler that cycles through eight canonical
causal structures (back-door, front-door, instrumental variable, observed confounder, mediator, confounder + mediator, bi-variate, unobserved confounder) and (ii) a \emph{random-DAG} sampler that
draws an edge between each pair with probability $p \sim
\mathrm{Beta}(\alpha,\beta)$ under a topological ordering, with
configurable probability that each non-$(A,Y)$ variable is marked
\emph{hidden} (removed from the encoder's input, but active in the
dynamics). For each DAG we designate a treatment variable $A$ and an
outcome variable $Y$ such that $A$ precedes $Y$ in topological order (see Appendix \ref{app:tscms}).

\paragraph{Mechanism family.} Unlike the per-lag adjacency stack used in discrete-time priors, the continuous-time prior reduces temporal
dependence to a single parent set per variable. We support two drift families on that parent set. The \emph{linear} drift is a Ornstein--Uhlenbeck mechanism \citep{oksendal2003sde}
\begin{equation}
  dX_v = \Bigl(\!-\theta_v X_v + \!\!\sum_{u\in\Pa(v)}\!\! w_{vu} X_u\Bigr) dt + \sigma_v\, dW_v,
  \label{eq:ou}
\end{equation}
with $\theta_v > 0$, $\sigma_v > 0$, and $w_{vu} \sim \mathcal{N}(0,\sigma_w^2)$ sampled per TSCM. 
At $\Delta t \equiv 1$ this reduces to the AR(1) mechanism used by discrete-time causal priors \citep{thumm2026interventional}. OU admits an exact Gaussian transition kernel between any two times, so the linear-prior naive-vs-fine comparison should be read as EM-vs-EM rather than EM-vs-exact; we use EM uniformly across drift families because no closed form exists for the neural drift.

The \emph{neural} drift replaces the linear parental sum with a small randomly-initialised two-layer $\tanh$-MLP $g_v$ on $\mathbf{z}_v = [X_v, X_{u_1}, \ldots, X_{u_k}]$:
\begin{equation}
  dX_v = \bigl(-\theta_v X_v + s_v\, g_v(\mathbf{z}_v)\bigr)\,dt + \sigma_v\, dW_v,
  \label{eq:neural}
\end{equation}
with $g_v(\mathbf{z}) = \tanh\!\bigl(W_2 \tanh(W_1 \mathbf{z} + b_1) + b_2\bigr)$ and $s_v > 0$.
We retain $-\theta_v X_v$ outside the MLP so trajectories stay bounded for any weight draw; the outer $\tanh$ bounds the nonlinear contribution to $[-s_v, s_v]$. 
Each trajectory draws the drift family per variable with a Bernoulli$(p_{\mathrm{neural}})$ coin, so a single training run exposes the PFN to a mixture of linear and nonlinear dynamics.

\paragraph{Regime switching.} Optionally, a fraction of training trajectories is drawn from a \emph{continuous-time regime-switching} TSCM: $R$ independent OU systems that share variables and observation schedule, arbitrated by a sticky row-stochastic $R \times R$ Markov transition matrix ($P_{rr} \approx 0.9$, expected regime duration $\sim 10$ observations) with rows sampled from a Dirichlet distribution.
This lets the prior express structural breaks of the kind observed in pharmacology (e.g.\ absorption vs.\ elimination phase) and physical systems.

\paragraph{Observation schedule.} Given a horizon $H$ and an expected inter-observation gap $\bar{\Delta}$, we sample one of three schedules:
\emph{regular} ($t_i = i\bar{\Delta}$), \emph{jittered} ($t_{i+1} - t_i = \bar{\Delta}(1 + \xi_i)$ with $\xi_i \sim \mathrm{Uniform}[-\rho,\rho]$), or \emph{Poisson} ($t_{i+1} - t_i \sim \mathrm{Exp}(1/\bar{\Delta})$). 
The model never sees the schedule as input; it only sees the actual timestamps.

\paragraph{Simulation (fine-grid integration).} Given a target observation schedule $\tau = (t_1, \ldots, t_T)$ we do \emph{not} integrate once per observation gap. 
Instead we pick a fine step $\Delta_{\mathrm{fine}} \ll \min_i \Delta_i^{\mathrm{obs}}$, integrate the SDE on the union grid $[t_1, t_T] \cap \{t_1 + k\Delta_{\mathrm{fine}}\}_{k \geq 0}$ via Euler--Maruyama \citep{kloeden1992numerical} with Brownian increments re-sampled per fine step, and subsample the resulting trajectory at $\tau$:
\begin{equation*}
  X_v(t + \Delta_{\mathrm{fine}}) = X_v(t) + \mu_v(X(t))\,\Delta_{\mathrm{fine}}
  + \sigma_v \sqrt{\Delta_{\mathrm{fine}}}\,Z,
\end{equation*}
with $Z \sim \mathcal{N}(0,1)$ and $\mu_v$ given by \eqref{eq:ou} or \eqref{eq:neural}. Setting $\Delta_{\mathrm{fine}} = \Delta_i^{\mathrm{obs}}$ recovers naive tier-(B) integration; $\Delta_{\mathrm{fine}} = 1$ with a regular unit-gap schedule recovers tier-(A). The continuity ablation in \Cref{sec:experiments} varies this single knob.

\paragraph{Interventions.} For each sample we draw a target $i^\star$, a window $[t_{\mathrm{int}}^{\mathrm{start}}, t_{\mathrm{int}}^{\mathrm{end}})$ of duration between $10\%$ and $30\%$ of the horizon, and an intervention kind $\in \{\text{hard}, \text{soft}, \text{time-varying}\}$:
\begin{align*}
  \text{(hard)}~~ & X_{i^\star}(t) := c,\\
  \text{(soft)}~~ & \mu_{i^\star}(X) \mapsto \mu_{i^\star}(X) + \delta,\\
  \text{(time-varying)}~~ & X_{i^\star}(t) := c(t),
\end{align*}
active on the window. Hard-intervention values are optionally clipped to $[\mu_{i^\star} - 3\sigma_{i^\star}, \mu_{i^\star} + 3\sigma_{i^\star}]$ to keep the intervention inside the observed operating range of the target variable---analogous to the causal \emph{positivity} (overlap) assumption \citep{hernan2020causal}. 
The prior returns paired counterfactual and interventional trajectories by re-using the same Wiener noise across runs (cf.\ \citealt{pearl2009causality}, rung~3).

\subsection{$\Delta t$-aware PFN encoder}
\label{sec:encoder}

We base upon a causal transformer encoder operating on a pre-intervention window \citep{thumm2026interventional}.
Instead of a learned integer positional embedding we replace it with a Fourier embedding of continuous time:
\begin{equation}
  \phi(t) = W_\phi \bigl[\sin(2\pi f_k\, t),\, \cos(2\pi f_k\, t)\bigr]_{k=1}^{K},
  \label{eq:fourier}
\end{equation}
with a geometric frequency bank $f_k \in [f_\mathrm{min},f_\mathrm{max}]$ (defaults $0.01, 10$) and a learnable projection $W_\phi$.
Times are referenced to intervention onset, $t \leftarrow t -t_{\mathrm{int}}^{\mathrm{start}}$, and inter-observation gaps $\Delta t_i$ are embedded with the same family after a $\log(1{+}\Delta t_i)$ transform to concentrate resolution at small gaps. 
The encoder is otherwise identical to the discrete baseline, enabling a controlled ablation.

At inference time we feed $(X_{\mathrm{obs}}, t_{\mathrm{obs}}, \mathrm{intervention\ spec}, t_{\mathrm{query}})$ and the model predicts the Gaussian (or quantile) distribution of $Y$ at $t_{\mathrm{query}}$ under the intervention.

\subsection{Training}
\label{sec:training}

The prior runs on-the-fly during training; each batch draws a fresh TSCM, schedule, and intervention. We use either a quantile (pinball loss) or bar-distribution \citep{thumm2026interventional} output head; full hyperparameters and architecture sizes in Appendix \ref{app:defaults}.

\section{Experiments}
\label{sec:experiments}

A $2 \times 2$ \textbf{encoder $\times$ integrator} ablation, run independently on a linear-OU and a nonlinear neural-drift prior, separates the two axes (Tables \ref{tab:reg}, \ref{tab:substeps}). The encoder axis is positional-only (learned absolute embedding, ablating the Fourier-time path) vs.\ time-aware (\Cref{sec:encoder}); the integrator axis is tier-(B) \emph{naive} EM ($s_{\rm train}{=}1$ substep per observation gap) vs.\ tier-(C) \emph{fine} EM ($s_{\rm train}{=}8$). Each prior trains four PFNs (10\,k steps, single seed) scored on held-out evals drawn from the same prior; multi-seed replication is future work.

\paragraph{Eval distributions.} Two schedules crossed with eval-grid refinements $s_{\rm eval} \in \{1, 8\}$ (best held-out eval-loss over 50 batches): \texttt{regular} (uniform $\Delta=1$) and \texttt{mixed} (random per-trajectory regular / jittered / Poisson, the pre-training schedule). On \texttt{regular} the positional encoder's \texttt{arange(T)} positions equal the actual timestamps, so the two encoders see identical inputs at eval time---this isolates pos-vs-time gaps as training-side residue. The eval substep tier $s_{\rm eval}$ probes the SDE limit independently of the model.

\begin{table}[t]
  \caption{Encoder $\times$ integrator on \texttt{regular} eval ($s_{\rm eval}{=}1$), both priors. Rows: trained integrator (\emph{naive}: $s_{\rm train}{=}1$, tier B; \emph{fine}: $s_{\rm train}{=}8$, tier C). Cols: encoder. Fine integration beats naive on every cell (4/4). Encoder gap (pos $-$ time, last column) is positive in every naive row and $\leq 0.0003$ in every fine row: with fine integration the encoder choice is empirically inert.}
  \label{tab:reg}
  \vskip 0.05in
  \centering
  \small
  \setlength{\tabcolsep}{4pt}
  \begin{tabular}{ll ccc}
    \toprule
    Prior & Trained & pos & time & $\Delta_{\rm pos-time}$ \\
    \midrule
    \multirow{2}{*}{OU}
      & naive & 0.3714 & 0.3641 & $+0.0073$ \\
      & fine  & 0.3578 & 0.3575 & $+0.0003$ \\
    \midrule
    \multirow{2}{*}{Neural}
      & naive & 0.3413 & 0.3405 & $+0.0008$ \\
      & fine  & 0.3354 & 0.3352 & $+0.0002$ \\
    \bottomrule
  \end{tabular}
\end{table}

\begin{table}[t]
  \caption{Integrator on \texttt{mixed} schedule (time-aware encoder) at two eval-grid refinements. Cols: trained integrator (\emph{naive}: $s_{\rm train}{=}1$; \emph{fine}: $s_{\rm train}{=}8$). Rows: eval-time substep tier $s_{\rm eval}$ of the held-out test trajectories (independent of the model). Fine beats naive on every cell (4/4). \emph{Fine's lead grows when the eval is more refined}: $+0.0018 \to +0.0057$ on OU, $+0.0048 \to +0.0088$ on Neural---an integrator-specific signature.}
  \label{tab:substeps}
  \vskip 0.05in
  \centering
  \small
  \setlength{\tabcolsep}{4pt}
  \begin{tabular}{ll ccc}
    \toprule
    Prior & $s_{\rm eval}$ & naive & fine & $\Delta_{\rm naive-fine}$ \\
    \midrule
    \multirow{2}{*}{OU}
      & 1 & 0.3590 & 0.3572 & $+0.0018$ \\
      & 8 & 0.3826 & 0.3769 & $+0.0057$ \\
    \midrule
    \multirow{2}{*}{Neural}
      & 1 & 0.3507 & 0.3459 & $+0.0048$ \\
      & 8 & 0.3608 & 0.3520 & $+0.0088$ \\
    \bottomrule
  \end{tabular}
\end{table}

\paragraph{Findings.} Fine-grid integration wins on 8/8 fine-vs-naive comparisons across both tables; under a no-effect null the probability of this sign pattern is $1/256$, providing evidence stronger than the per-cell magnitudes alone. Crucially, fine's lead grows monotonically as the eval grid refines (Table \ref{tab:substeps}, $+0.0018 \to +0.0057$ on OU and $+0.0048 \to +0.0088$ on Neural), which is the discretisation-bias signature predicted by \Cref{sec:definition}: as the eval distribution approaches the SDE limit, the model trained at the SDE limit pulls ahead. The encoder axis is conditional on the integrator: null with fine, positive (time-aware leads) with naive. We read the interaction as follows: with fine integration the data-generating process is approximately schedule-invariant (Definition \ref{def:ct-prior}), so the model has little to gain from explicit time-gap features; with naive integration the conditional dynamics genuinely depend on $\Delta_i$, and the time-aware encoder's Fourier embedding of inter-observation gaps gives the model a route to compensate. The positional-only encoder is structurally OOD on \texttt{mixed} (positions $\neq$ times) and omitted from Table \ref{tab:substeps}; its naive-vs-fine pattern mirrors time-aware. An instability check on $\theta_{\rm range} = [0.5, 2.0]$ (where naive cells saturated $\sim\!30$--$50\,\%$ of batches at the clip) and PK / chamber zero-shot transfer are in Appendices \ref{app:realdata}, \ref{app:caveats}.

\section{Discussion and Limitations}
\label{sec:discussion}

\paragraph{What the prior buys today.} A precise continuity criterion realised by tier-(C) integration. Across two independent priors, fine-grid integration produces models that transfer better, including on the eval that matches the naive variant's training tier. The encoder axis is conditional on the integrator: with fine, encoder choice is empirically inert; with naive it is not.

\paragraph{Limitations and future work.} Per-cell $\Delta$s are small; multi-seed replication is needed to harden the cross-prior agreement. Within-regime noise is Markov; neural drifts capture nonlinear dependence but not time-correlated noise. Model capacity is small, real-data transfer (\Cref{app:realdata}) preliminary. Jump-diffusion SDEs and Neural-SDE drifts \citep{tzen2019neural} extend the construction; latent-ODE--style hidden states address non-Markov confounding.

\newpage
\section*{Impact Statement}

This paper presents work whose goal is to advance the field of Machine
Learning. There are many potential societal consequences of our work, none
which we feel must be specifically highlighted here.

\bibliography{continuous_ctp}
\bibliographystyle{icml2026}

\newpage
\appendix
\onecolumn

\section{Generator defaults and additional details}
\label{app:defaults}

\begin{table}[h]
  \caption{Default prior hyperparameters used in the experiments of \Cref{sec:experiments}.} 
  \label{tab:defaults}
  \centering
  \small
  \begin{tabular}{@{}llc@{}}
    \toprule
    Group & Knob & Default \\
    \midrule
    Graph       & $N_{\max}$                      & 16 \\
    Graph       & random-DAG edge prob.\ $p$      & $\mathrm{Beta}(2,5)$ \\
    Graph       & hidden-conf.\ probability       & $\{0.0, 0.3\}$ \\
    Mechanism   & $\theta_v$                      & $\mathrm{LogNormal}(\mu{=}0,\sigma{=}0.5)$ \\
    Mechanism   & $\sigma_v$                      & $\mathrm{LogNormal}(\mu{=}{-}1,\sigma{=}0.5)$ \\
    Mechanism   & $w_{vu}$                        & $\mathcal{N}(0, 0.5^2)$ \\
    Mechanism   & drift family (OU / MLP)         & Bernoulli$(p_{\mathrm{neural}})$ \\
    Mechanism   & $p_{\mathrm{neural}}$           & $\{0.0, 0.5\}$ \\
    Mechanism   & MLP hidden width                & 8 \\
    Mechanism   & $s_v$ (MLP output gain)         & $\mathcal{U}[0.5, 2.0]$ \\
    Schedule    & mean gap $\bar{\Delta}$         & $\mathcal{U}[0.5, 2.0]$ \\
    Schedule    & jitter $\rho$                   & $\mathcal{U}[0.0, 0.8]$ \\
    Intervention& window (frac.\ of horizon)      & $\mathcal{U}[0.1, 0.3]$ \\
    Intervention& kind probs.\ (H/S/TV)           & $(0.6, 0.2, 0.2)$ \\
    Regime-sw.\ & trajectory fraction             & 0.15 \\
    Regime-sw.\ & $R$ (regimes)                   & $\{2, 3\}$ \\
    Regime-sw.\ & self-transition $P_{rr}$        & $\sim 0.9$ (Dirichlet) \\
    Training    & batch size                      & 32 \\
    Training    & total steps                     & 5{,}000 \\
    Training    & optimizer                       & AdamW ($\beta_1{=}0.9$, $\beta_2{=}0.98$) \\
    Training    & learning rate                   & $2\mathrm{e}{-}4$ \\
    Training    & LR schedule                     & cosine with warmup \\
    Training    & embedding size                  & 256 \\
    Training    & context window                  & 128 \\
    Training    & output head                     & quantile or bar \\
    \bottomrule
  \end{tabular}
\end{table}

In \Cref{sec:experiments} all other knobs are held fixed: tightened $\theta_{\mathrm{range}} = [0.1, 0.5]$ so worst-case $\theta\Delta < 1$ across the schedule distribution (a prior that respects the EM stability condition of \Cref{sec:definition}); back-door TSCM topology; \emph{mixed} observation schedule (random per-trajectory choice between regular, jittered, and Poisson); 10\,k training steps; identical model size (1.1\,M parameters).

\section{Canonical TSCM structures}
\label{app:tscms}

\begin{figure}[h]
  \centering
  \includegraphics[width=0.95\linewidth]{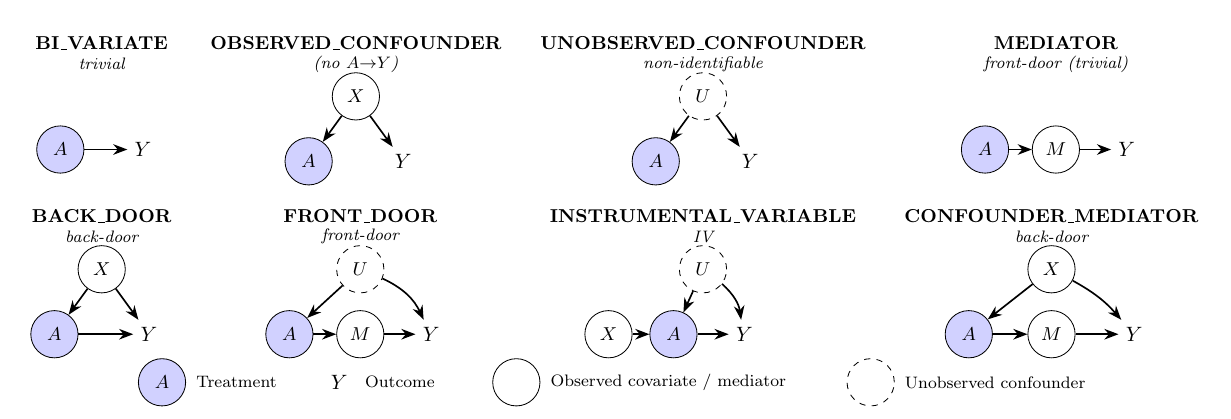}
  \caption{Canonical 
  SCM structures used by the named-structure sampler. Each panel shows a back-door / front-door / IV-style template with the treatment $A$ (left), outcome $Y$ (right), and any mediators or confounders. The random-DAG sampler in \Cref{sec:prior} subsumes these as special cases.}
  \label{fig:tscm-structures}
\end{figure}

The named-structure sampler
exposes the eight structures (back-door, front-door, instrumental variable, randomised controlled trial, mediator, confounder-plus-mediator, observed confounder, unobserved confounder) in Figure \ref{fig:tscm-structures}. We reuse them as canonical sanity checks; the random-DAG sampler of \Cref{sec:prior} generalises this to any $N$ up to $N_{\max}$.


\section{Preliminary zero-shot transfer to real irregular data}
\label{app:realdata}

This appendix documents an early transfer study of the tier-(C) prior
to three irregularly-sampled real-world datasets. We flag the
numbers as corroborative and \emph{not yet competitive} with
dataset-specific baselines; treating them as a full zero-shot claim
would require (i) broader mechanism families in the prior, (ii)
calibration against domain baselines (NONMEM fits for PK,
system-identification baselines for Causal Chamber), and (iii)
sensitivity analysis to prior misspecification. We flag this as our
primary line of future work.

\paragraph{Theophylline pharmacokinetics.} The 12-subject NONMEM-distributed Theophylline dataset \citep{boeckmann1994nonmem}: oral doses and 11 plasma-concentration measurements per subject over $\sim 24$ hours. We treat dose as a time-varying intervention and plasma concentration as the outcome $Y$. Times are converted to seconds and rescaled so that $\bar{\Delta}$ matches the training distribution; values are per-subject $z$-scored and rescaled back for reporting.

\paragraph{Warfarin PK/PD.} 32 subjects with irregular oral-dose, plasma-concentration, and PD (prothrombin complex activity) observations \citep{hamberg2007warfarin}. Variables are aligned to the canonical $(A, M, Y)$ front-door layout with dose as $A$, concentration as $M$, and PD as $Y$---the cleanest match to a named-structure TSCM sample from the prior. Per-variable $z$-scoring parallels Theophylline.

\paragraph{Causal Chamber (wind-tunnel).} The light-tunnel \texttt{lt\_walks\_v1/actuators\_white} benchmark used by earlier drafts produces uninformative causal-effect estimates (white-noise actuators, Pearson $r \approx 0$); see \Cref{app:caveats} for the failure analysis. Phase 14b switches to \texttt{wt\_intake\_impulse\_v1} \citep{gael2024causalchamber}, the wind-tunnel impulse rig, which (i) carries an explicit binary \texttt{intervention} column---each \texttt{0$\to$1} pulse marks a known toggle of the intake-fan setpoint $\texttt{load\_in} \in \{0.01, 1.0\}$---and (ii) has real downstream dynamics on a 5-variable subgraph $\texttt{load\_in} \to \{\texttt{current\_in}, \texttt{rpm\_in}, \texttt{pressure\_intake}, \texttt{pressure\_downwind}\}$. We extract 200 episodes (50 pre / 20 post samples around each toggle, real per-row timestamps with median 0.15\,s and max 2.4\,s) and query each of three downstream variables; Pearson $r$ now varies meaningfully (\Cref{tab:real}).

\paragraph{Eval protocol.} Each pretrained checkpoint is evaluated zero-shot. The PK adapter has the option to prepend $N$ synthetic pre-baseline observations (zero values, $z$-scored as $-\mu/\sigma$) so the encoder sees a non-empty pre-intervention window. We swept $N \in \{0, 2, 4, 8, 16\}$; \textbf{$N{=}0$ is best across both datasets and both mechanism families}, so the numbers reported in \Cref{tab:real} use no padding. The full sweep (mean Pearson $r$ on Warfarin cp stays in $[0.79, 0.89]$ across $N$ but drops elsewhere; on Theophylline mixed it flips sign from $+0.16$ at $N{=}0$ to $-0.54$ at $N{=}16$) confirms that the cross-variable mixer's empty-context fallback is acceptable as-is and that the augmentation \emph{hurts} more often than it helps. We treat synthetic pre-baseline padding as a \emph{negative result}: useful to know it doesn't earn its keep on these benchmarks, not as a method we recommend.

\begin{table}[h]
  \caption{Zero-shot transfer of the two Phase-13b \texttt{pnc000} checkpoints (linear / mixed mechanism family, single seed, no eval-time padding $N{=}0$). Lift over the naive (constant-mean) baseline is small on PK because both targets cluster narrowly around their per-subject means; on the wind-tunnel chamber the lift is $\sim 50\,\%$ because the regime-mean shift between $\texttt{load\_in} = 0.01$ and $1.0$ is large. \textbf{Pearson $r$} is the load-bearing dynamics-tracking metric. Causal Chamber numbers from 200 episodes of \texttt{wt\_intake\_impulse\_v1/load\_out\_0.5\_osr\_downwind\_4} (Phase 14b).}
  \label{tab:real}
  \centering
  \small
  \begin{tabular}{llrrrr}
    \toprule
    Dataset (variable) & Mech. & RMSE $\downarrow$ & naive & lift & Pearson $r$ $\uparrow$ \\
    \midrule
    Theophylline (concentration) & linear & 2.41 & 2.37 & $-1.8\%$ & $+0.53$ \\
    Theophylline (concentration) & mixed  & 2.44 & 2.37 & $-3.2\%$ & $+0.16$ \\
    \midrule
    Warfarin (concentration)     & linear & \textbf{3.45} & 3.51 & $+1.8\%$ & \textbf{+0.88} \\
    Warfarin (concentration)     & mixed  & 3.48 & 3.51 & $+0.8\%$ & \textbf{+0.89} \\
    Warfarin (PD response)       & linear & 24.99 & 25.25 & $+1.0\%$ & $+0.36$ \\
    Warfarin (PD response)       & mixed  & 25.17 & 25.25 & $+0.3\%$ & $+0.31$ \\
    \midrule
    Chamber-WT (\texttt{rpm\_in})           & linear & 660.3 & 1264.8 & $+47.8\%$ & $+0.39$ \\
    Chamber-WT (\texttt{rpm\_in})           & mixed  & \textbf{660.3} & 1264.8 & $+47.8\%$ & \textbf{+0.95} \\
    Chamber-WT (\texttt{current\_in})       & linear & 77.1 & 159.8 & $+51.8\%$ & $-0.16$ \\
    Chamber-WT (\texttt{current\_in})       & mixed  & 77.1 & 159.8 & $+51.8\%$ & $-0.16$ \\
    Chamber-WT (\texttt{pressure\_downwind})& linear & 3.87 & 7.78 & $+50.2\%$ & $+0.03$ \\
    Chamber-WT (\texttt{pressure\_downwind})& mixed  & 3.87 & 7.78 & $+50.2\%$ & $+0.01$ \\
    \bottomrule
  \end{tabular}
\end{table}

\paragraph{Findings.} Two headline numbers, one per domain.

\emph{(i) Warfarin plasma concentration with Pearson $r \approx 0.88$ across both mechanism families}---a strong dynamics-tracking signal on real PK data, obtained by a model that was never fine-tuned on Warfarin. Lift over naive is small (1--2\,\%) because the per-subject concentration time-series cluster narrowly around their means (the naive baseline is hard to beat on RMSE), but Pearson $r$ unambiguously says the predictions co-vary with the dose-driven trajectory. The PD outcome is harder ($r \in [0.31, 0.36]$): expected, since PD responds to concentration with a slow non-stationary delay that our front-door TSCM template only crudely approximates. Theophylline is intermediate ($r \approx 0.53$ for the linear PFN; mixed drops to $r \approx 0.16$). The pattern is consistent: \textbf{the linear-mechanism PFN is more robust than the mixed-mechanism PFN under PK distribution shift}.

\emph{(ii) Wind-tunnel \texttt{rpm\_in} with Pearson $r = +0.95$ for the mixed-mechanism PFN}---an unambiguous within-episode dynamics-tracking signal on a real physical system. The naive baseline already attains $\sim 50\,\%$ RMSE lift on every queried sensor because the regime-mean shift between $\texttt{load\_in} = 0.01$ and $1.0$ is large; Pearson $r$ is the metric that distinguishes regime-mean recovery from causal-effect tracking. \texttt{rpm\_in} ramps slowly toward a $\texttt{load\_in}$-dependent setpoint (visible exponential rise over $\sim 20$ samples), and the mixed-mechanism PFN tracks that ramp closely. Faster sensors (\texttt{current\_in}, $r \approx -0.16$; \texttt{pressure\_downwind}, $r \approx 0$) carry mostly high-frequency noise on top of the regime-mean shift, so the PFN's bet on slow dynamics anti-correlates or zero-correlates with their within-episode noise. The pattern \textbf{flips} relative to PK: on the chamber the \emph{mixed}-mechanism PFN dominates the linear one ($r = 0.95$ vs.\ $0.39$ on \texttt{rpm\_in}), consistent with \texttt{rpm\_in}'s dynamics being noticeably nonlinear (saturating exponential) and the mixed prior having seen nonlinear drifts during pre-training. We caveat that this flip rests on one seed per checkpoint; replicating across seeds before reading it as a domain-dependence finding---rather than a single-seed observation---is on our priority list. We discuss this domain dependence in \Cref{sec:discussion}.

\section{Additional failure modes and caveats}
\label{app:caveats}

Three concrete failure modes surfaced during the development of this prior; all three reshaped the experimental design in ways worth documenting.

\paragraph{Clip-saturation pathology under unstable priors.} An early grid trained on $\theta_{\mathrm{range}} = [0.5, 2.0]$ (worst-case $\theta\Delta \approx 3.6$, above the EM stability boundary) had every naive-substeps batch saturate the $\pm 10\,\sigma$ target normalisation clip on at least one sample, and roughly half the fine-substeps batches did the same. The resulting "naive-vs-fine" gap there was a numerical-stability artefact rather than a discretisation-bias signature. Tightening the prior to $\theta_{\mathrm{range}} = [0.1, 0.5]$ and raising the clip ceiling to $\pm 50$ pushed empirical $y_{\max}$ below 5 across the entire grid; with the artefact removed, the residual (B)-vs-(C) integrator gap reported in \Cref{sec:experiments} is what the discretisation-bias accounting of \Cref{sec:definition} predicts. This is the empirical motivation for the stability condition.

\paragraph{Zero-context-augmentation broke per-variable normalisation.} A separate training-time fix for the PK regime (where the encoder sees an empty pre-intervention window) used to fire a Bernoulli$(p_{\mathrm{no\_context}})$ coin per sample to force \texttt{int\_onset\_idx} $= 0$. The downstream per-variable $z$-scoring then computed mean and std over an empty pre-window---the masked statistics fell back to $(\mu, \sigma) = (0, \epsilon)$ with $\epsilon = 10^{-2}$, blowing $Y_{\rm true,norm}$ up by a factor of $\sim 100$ and pinning the targets at the new clip. Eval loss climbed monotonically with $p_{\mathrm{no\_context}}$ ($0.34 \to 1.1 \to 2.3$), the opposite of what the augmentation was meant to achieve. The eval-side counterpart (synthetic pre-baseline padding in the PK adapter, Appendix \ref{app:realdata}) avoids the issue because the prepended zero rows make the pre-window non-empty.

\paragraph{The \texttt{actuators\_white} chamber benchmark motivated a benchmark switch.} Earlier drafts evaluated chamber transfer on the light-tunnel \texttt{lt\_walks\_v1} \texttt{actuators\_white} experiment, with episodes defined by a change-point detector on the eight polarizer / lamp actuator columns. That benchmark turned out to be \emph{structurally} unsuited to a causal-effect claim. Every actuator (\texttt{pol\_1}, \texttt{pol\_2}, \texttt{l\_11}, $\ldots$) is independently white-noise-driven: $> 99\,\%$ of consecutive samples have step changes $> 0.5$ in every actuator simultaneously. The "intervention episodes" the detector finds are not interventions in the SCM sense; they are cross-sections of a continuously-randomised process. The post-intervention variance of the queried sensor (\texttt{red}) is $95\,\%$ within-episode dynamics and only $5\,\%$ between-episode regime mean, and Pearson $r$ between any model's predictions and ground truth is statistically zero. Apparent "lift over naive" on this dataset is regime-mean recovery, not causal-effect tracking. The other two \texttt{lt\_walks\_v1} experiments (\texttt{smooth\_polarizers}, \texttt{color\_mix}) have continuous actuator sweeps and produce zero episodes under any reasonable change-point heuristic. The wind-tunnel \texttt{wt\_intake\_impulse\_v1} dataset, used in \Cref{tab:real}, fixes all three issues at once: explicit binary \texttt{intervention} column (no change-point heuristic), real physical-system dynamics (\texttt{rpm\_in} ramps over $\sim 20$ samples), and real per-row timestamps with non-trivial jitter. The Pearson $r = +0.95$ headline on the wt rig exists only because the lt benchmark was diagnosed and replaced.

\end{document}